\documentclass[sigconf]{acmart}
\AtBeginDocument{%
  }

\copyrightyear{2026}
\acmYear{2026}
\setcopyright{cc}
\setcctype{by}
\acmConference[CAIN '26]{2026 IEEE/ACM 5th International Conference on AI Engineering - Software Engineering for AI}{April 12--13, 2026}{Rio de Janeiro, Brazil}
\acmBooktitle{2026 IEEE/ACM 5th International Conference on AI Engineering - Software Engineering for AI (CAIN '26), April 12--13, 2026, Rio de Janeiro, Brazil}
\acmDOI{10.1145/3793653.3793767}
\acmISBN{979-8-4007-2475-6/2026/04}

\usepackage{subcaption}
\usepackage{algorithm2e}

\begin{document}

\title{Optimising for Energy Efficiency and Performance in Machine Learning}

\author{Emile Dos Santos Ferreira}
\orcid{0009-0003-2155-5435}
\affiliation{%
  \institution{University of Cambridge}
  \city{Cambridge}
  \state{England}
  \country{UK}
}
\email{edsf2@cantab.ac.uk}

\author{Andrei Paleyes}
\orcid{0000-0002-3703-8163}
\affiliation{%
  \institution{University of Cambridge}
  \city{Cambridge}
  \state{England}
  \country{UK}
}
\affiliation{%
  \institution{Pasteur Labs}
  \city{Brooklyn}
  \state{NY}
  \country{USA}
}
\email{ap2169@cl.cam.ac.uk}

\author{Neil D. Lawrence}
\orcid{0000-0001-9258-1030}
\affiliation{%
  \institution{University of Cambridge}
  \city{Cambridge}
  \state{England}
  \country{UK}
}
\email{ndl21@cam.ac.uk}

\renewcommand{\shortauthors}{Ferreira, et al.}

\begin{abstract}
    The ubiquity of machine learning (ML) and the demand for ever-larger models bring an increase in energy consumption and environmental impact.
    However, little is known about the energy scaling laws in ML, and existing research focuses on training cost---ignoring the larger cost of inference.
    Furthermore, tools for measuring the energy consumption of ML do not provide actionable feedback.
    
    To address these gaps, we developed Energy Consumption Optimiser (ECOpt): a hyperparameter tuner that optimises for energy efficiency and model performance.
    ECOpt quantifies the trade-off between these metrics as an interpretable Pareto frontier. This enables ML practitioners to make informed decisions about energy cost and environmental impact, while maximising the benefit of their models and complying with new regulations.
    
    Using ECOpt, we show that parameter and floating-point operation counts can be unreliable proxies for energy consumption, and observe that the energy efficiency of Transformer models for text generation is relatively consistent across hardware.
    These findings motivate measuring and publishing the energy metrics of ML models.
    We further show that ECOpt can have a net positive environmental impact and use it to uncover seven models for CIFAR-10 that improve upon the state of the art, when considering accuracy and energy efficiency together.
\end{abstract}


%

\maketitle

\section{Introduction}
The exponential growth in computational complexity of machine learning (ML) models~\cite{Desislavov2023} drives an increase in energy cost and environmental impact. Recent examples include Stable Diffusion XL, which consumes approximately a phone charge's worth of energy to generate an image~\cite{Luccioni2024powerhungry}, and GPT-4, which used more than \$5 million of electricity during training~\cite{cottier2025rising}.
The carbon footprint of ML is one of the pressing concerns of society~\cite{qiu2023first}, with training BLOOM~\cite{le2023bloom} emitting an estimated 24.7 tonnes of CO$_2$eq~\cite{luccioni2023estimating}.
This concern continues to grow.
The cost of frontier model\footnote{Defined as a model that is amongst the 10 most computationally expensive models at the time of its release.} training has been doubling each year since 2016 and is projected to reach \$1 billion per model by 2027~\cite{cottier2025rising}.
In light of this, a responsible approach to model deployment should include a thorough understanding of the energy cost and environmental impact.
Indeed, recent regulation, including Article 53 of the European Union (EU) Artificial Intelligence (AI) Act~\cite{EUAIAct2024}, requires model providers to publish their energy usage.

However, ML research is focused on improving model performance\footnote{We define performance as how well a model accomplishes its intended task, be it measured by accuracy, mean squared error or some other metric.}, and models are rarely published with energy consumption metrics.
We find that the cost metrics which are frequently published -- parameter and floating-point operation (FLOP) counts -- can be uncorrelated with energy usage.
Furthermore, existing research on the environmental impact of ML mostly focuses on training cost, ignoring inference~\cite{chitakunye2025mlcomposition, nguyen2025on-deviceorremote}.
This is despite the fact that the cost of inference can far exceed that of training~\cite{Desislavov2023}; Meta reserves 70\% of its ML data centres for inference~\cite{MLSYS2022_462211f6} and Amazon Web Services states that inference can account for up to 90\% of the cost of ML compute~\cite{aws}.

At present, little is known about energy scaling laws in ML---the empirical relationships amongst energy efficiency, model size and model performance.
Existing solutions do not fully address this gap.
Tools for estimating the carbon footprint of models~\cite{Bouza2023, anthony2020, Igescu2023} require high code coupling and do not provide actionable feedback on how to improve energy efficiency.
Hyperparameter tuners~\cite{fetterman2023tunescalehyperparameteroptimization, you2023zeus} do not account for environmental impact or consider inference.
Model inference systems~\cite{salmani2023} do not generally use energy cost metrics, nor do they support model development.

To address the aforementioned shortcomings, we have developed Energy Consumption Optimiser (ECOpt): a Python framework that simultaneously optimises model hyperparameters for energy efficiency and performance.
ECOpt quantifies the trade-off between these metrics for any ML model as an interpretable Pareto frontier---discovered using multi-objective Bayesian optimisation.
ML practitioners can use this frontier to make informed decisions about the energy cost and environmental impact of deploying their models, while maximising performance.
ECOpt can be used to support the development of new models and to optimise existing deployments. Our thesis is that, by optimising hyperparameters for both energy efficiency and performance, we can reduce the energy cost of ML without sacrificing the quality of inference.

In this work, we put forth the following contributions:
\begin{itemize}
    \item We formulate the search for the trade-off between performance and energy efficiency in ML as a multi-objective optimisation task.
    \item We describe a fully automated procedure, based on multi-objective Bayesian optimisation, for discovering an empirical Pareto frontier characterising said trade-off.
    \item We perform an extensive experimental evaluation of our proposed method and highlight important results. These include showing that the energy efficiency of Transformer models is relatively consistent across hardware, uncovering an energy scaling law for these models, and demonstrating that our method can have a positive environmental impact. 
    \item We provide an open-source implementation of ECOpt as a reusable Python framework\footnote{\url{https://github.com/emileferreira/ecopt}}. This includes the experiment scripts for reproducibility.
\end{itemize}



\section{Related work}
\label{sec:related}

Related work in the area of efficient deep learning explores various approaches to balancing model performance with energy consumption.
\citet{2025Aquinoindex} propose an energy consumption index for deep learning models, quantifying the compromise between model performance and energy consumption using the Jappa-Energy metric. However, this approach is limited to classification tasks and, critically, reduces these objectives to a single efficiency metric, potentially overlooking valuable nuances. We advocate for a more granular approach: measuring and optimising both energy efficiency and performance independently. Zeus~\cite{you2023zeus} and Perseus~\cite{chung2024reducingenergybloat} form a family of tools for optimising the energy consumption of recurring deep neural network training. They are characterised by a focus on optimising the batch size and hardware energy limit. In contrast, ECOpt optimises any hyperparameters for both training and inference.

Hyperparameter tuning can lead to marked improvements in model performance and efficiency---including energy consumption \cite{Gutierrez2022, yarally2023uncovering}. Bayesian optimisation (BO) is widely used for hyperparameter tuning and is the technique that underpins the growing field of automated ML (AutoML)~\cite{Karmaker}.
\citet{yarally2023uncovering} treat energy consumption and accuracy as metrics of equal importance in hyperparameter tuning, but only consider training energy cost and do not provide a reusable tool.
\citet{fetterman2023tunescalehyperparameteroptimization} use BO in Cost-Aware Pareto Region Bayesian Search (CARBS) to search for hyperparameters which balance model accuracy and training time.
However, CARBS does not measure energy consumption, consider environmental impact or  support choice hyperparameters.
Furthermore, it ignores inference efficiency and requires the user to instrument their training code with BO steps---shortcomings that our work addresses directly.

Neural architecture search (NAS) is an AutoML technique used to uncover an optimal model architecture within a defined search space.
Since the model architecture can be configured using hyperparameters, NAS can be seen as a form of hyperparameter optimisation.
\citet{Lu2021} attempt to reduce the energy consumption of NAS while \citet{eriksson2021latency} perform NAS for on-device language models under the constraint of latency.
ECOpt supports NAS for arbitrary ML models and extends the constraints to include energy and carbon emissions.
We perform NAS using ECOpt in Section~\ref{sec:exp-nas}.

Instead of searching for an optimal architecture under fixed conditions, \citet{salmani2023} propose an ML inference system that dynamically selects models based on an objective function that encompasses accuracy, cost in terms of required central processing unit (CPU) cores, and latency.
This adaptive selection process is complementary to ECOpt's goal of identifying the most energy-efficient model at each level of accuracy and suggests potential synergy between our methodologies.

\section{Motivation}

Our work proposes to quantify the trade-off between performance and energy efficiency through repeated evaluations of a model with different hyperparameter values. It may seem that the energy required for this would negate any potential savings.
However, most of the ML workload is spent on inference tasks~\cite{MLSYS2022_462211f6, aws}, and improper hyperparameters can lead to markedly higher energy usage---even when obtaining similar model performance~\cite{geissler2024power}. For example, quantisation can increase inference energy efficiency by up to 70\% without decreasing performance~\cite{tschand2025mlperfpower}.

ECOpt optimises models for these inference tasks by identifying hyperparameters that maximise energy efficiency and performance, thereby minimising energy expenditure. Furthermore, we use multi-objective Bayesian optimisation as the optimisation method because of its sample efficiency, thus reducing the number of steps required for optimisation~\cite{yarally2023uncovering}.

As an example use case, an ML practitioner might be tasked with developing a recommender system that will be used to make millions of inferences per day. The system must achieve at least 90\% user satisfaction, based on a validation dataset of reviews. ECOpt automates the process of optimising the hyperparameters of the system for both inference energy efficiency and for user satisfaction. This automation leaves the practitioner free to focus on other tasks. Once the optimisation experiment is complete, they could then choose the most energy-efficient configuration that achieves 90\% user satisfaction---ignoring configurations with equal performance but lower efficiency.

The efficiency of the recommender system will lead to significant cost savings over its deployment lifetime.
Once deployed, the system could be further optimised by ECOpt to fully utilise the deployment hardware and reduce energy cost.
We demonstrate such savings in Section~\ref{sec:exp-batch}, and present the energy consumption and environmental impact of our work in Section~\ref{sec:exp-setup}.

\section{Methodology}
\label{sec:implementation}

\subsection{Overview}

Our primary goal is to quantify the trade-off between the performance and energy consumption of an ML model. We formulate the discovery of this trade-off as an optimisation problem with two objectives. In this section, we provide a detailed description of our optimisation process.

Optimisation of the chosen ML model is conducted with respect to its hyperparameters $\boldsymbol{h}$, which together form a search space that we denote as $S$. The dimensionality of $S$ is defined by the number of hyperparameters, and the cardinality of $S$ is determined by their domains. Let $S_i$ represent the domain of hyperparameter $h_i \in \boldsymbol{h}$ for $i = 1, 2, \ldots, n$. The search space is defined as the Cartesian product of the domains of the hyperparameters: $S = S_1 \times S_2 \times \cdots \times S_{n}$. Let the functions $f_p: S \to \mathbb{R}$ and $f_e: S \to \mathbb{R}$ map from the hyperparameter space $S$ to model performance and energy efficiency, respectively. Our aim is to find the Pareto set $\mathcal{P}(S)$ for the two-dimensional objective function $\boldsymbol{f} = [f_p, f_e]$.

Given a model implementation $\mathcal{I}$ and a dataset $\mathcal{D}$, the mapping of a point $\boldsymbol{h} \in S$ to the objective space proceeds as follows. First, the dataset $\mathcal{D}$ is divided into $\mathcal{D}_{train}$, for model training, and $\mathcal{D}_{eval}$, for inference evaluation. This separation is maintained for all optimisation steps. We then construct and train the model $\mathcal{M}$ using the hyperparameters defined by $\boldsymbol{h}$, resulting in a trained model $\mathcal{M}'$. While ECOpt can optimise any hyperparameters, for the purposes of the experiments reported in this paper, the training parameters (such as learning rate and number of epochs) are fixed across optimisation steps. Finally, we conduct inference with $\mathcal{M}'$ on $\mathcal{D}_{eval}$, measuring $\boldsymbol{f}(\boldsymbol{h})$ according to the chosen metrics. This process is formalised in Algorithm~\ref{algo:optimisation}.

\begin{algorithm}
\caption{The ECOpt optimisation procedure. If the model $\mathcal{M}$ has already been trained, the training call can be optionally skipped.}
\label{algo:optimisation}
\KwIn{Model implementation $\mathcal{I}$, dataset $\mathcal{D}$, maximum iterations $T$, two-dimensional objective function $\boldsymbol{f}$}
\KwOut{Pareto frontier $\mathcal{P}$}
\BlankLine
$\mathcal{D}_{\text{train}}, \mathcal{D}_{\text{eval}} \gets \text{Split}(\mathcal{D})$\;
$\mathcal{P} \gets \emptyset$\;
\For{$t = 1$ \KwTo $T$}{
    $\boldsymbol{h}_t \gets \text{Sample}(t)$\;
    $\mathcal{M} \gets \text{Construct}(\mathcal{I}, \boldsymbol{h}_t)$\;
    $\mathcal{M}' \gets \text{Train}(\mathcal{M}, \mathcal{D}_{\text{train}}, \boldsymbol{h}_t)$\;
    $\mathcal{P} \gets \text{Update}(\mathcal{P}, \boldsymbol{f}(\mathcal{M}', \mathcal{D}_{\text{eval}}, \boldsymbol{h}_t))$\;
}
\Return{$\mathcal{P}$}\;
\BlankLine
\end{algorithm}

The choice of $f_p$ depends on the task and model. For example, regression problems might use mean squared error and classification problems could use accuracy or F1 score. We leave the implementation $f_p$ open to the user in the model wrapper class. It typically involves applying the model to $\mathcal{D}_{eval}$ and comparing the model outputs to the expected outputs. In Section \ref{sec:experiments}, we declare the chosen measure of $f_p$ for each experiment. Meanwhile, the choice of $f_e$ is more involved, and we discuss it below in Section~\ref{section:energy-measure}.

\begin{figure}
    \begin{center}
        \includegraphics[width=0.95\columnwidth]{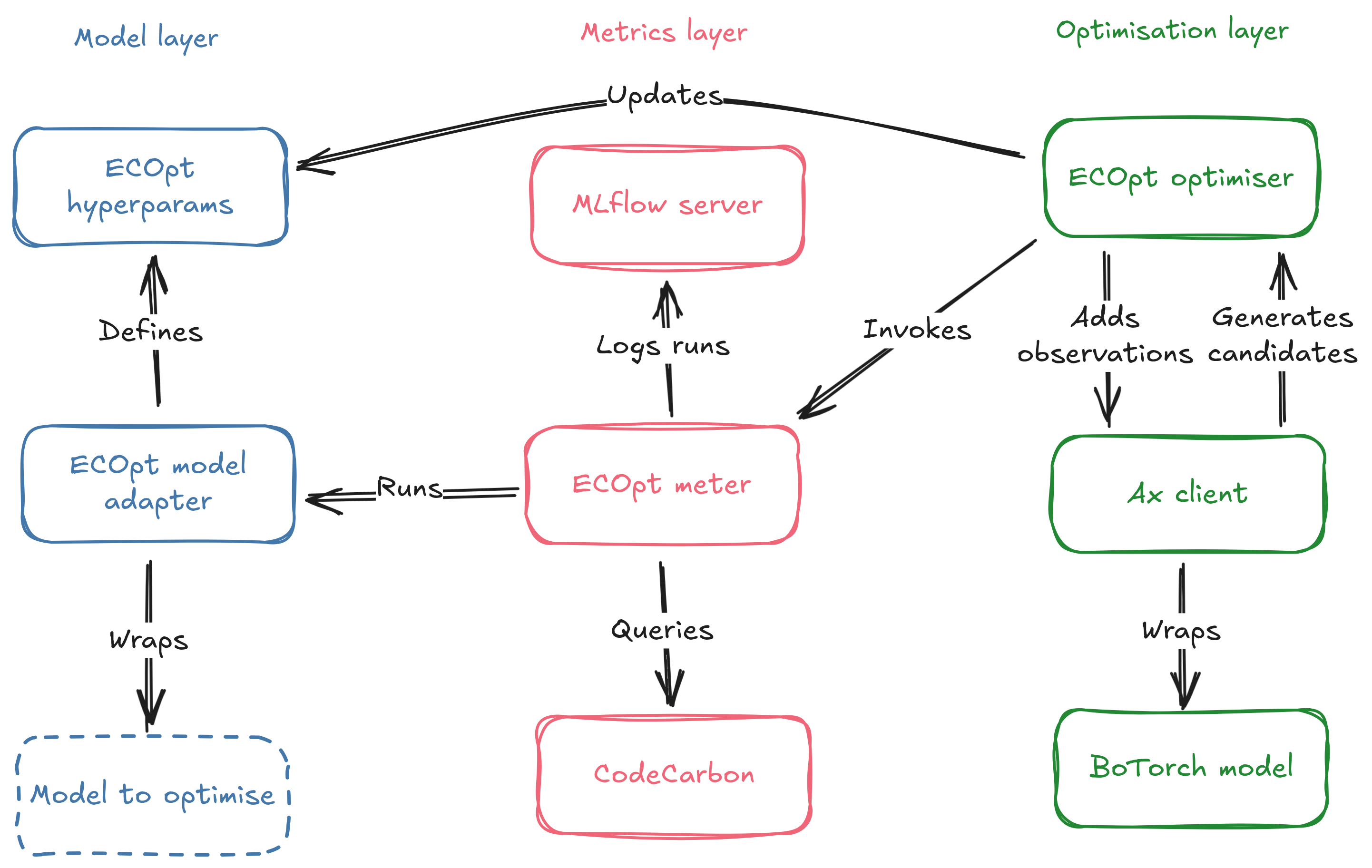}
    \end{center}
    \caption{An overview of the ECOpt system. The ECOpt meter can be used independently of the optimiser, to measure energy efficiency and performance metrics. The ECOpt optimiser performs multi-objective optimisation of the model hyperparameters.}
    \label{fig:ecopt-design}
    \Description{The ECOpt model adapter wraps an arbitrary machine learning model to optimise and defines ECOpt hyperparameters. These are updated by the ECOpt optimiser, which invokes the ECOpt meter. The meter runs the model adapter while querying CodeCarbon and then logs the runs to an MLflow server. The optimiser adds its oberservations to an Ax client, which wraps a BoTorch model, and receives candidate hyperparameter configurations in response.}
\end{figure}

ECOpt is implemented as a Python package and tested by a suite of unit tests in the source code provided. It is designed to maximise reuse, allowing for plug-and-play experience with respect to models, metrics and meters. We depict a high-level overview of the software components of ECOpt in Figure~\ref{fig:ecopt-design}.

\subsection{Measuring energy efficiency}\label{section:energy-measure}

We now consider how to measure energy efficiency $f_e$.
Measuring the energy consumption of ML models poses distinct challenges; models are executed on various hardware platforms ranging from general processors to custom accelerators~\cite{lim2021f1}, they require high data parallelism and they are often deployed on distributed systems. Moreover, energy consumption is sometimes affected by factors other than compute. The global average power usage effectiveness (PUE) of data centres is $1.56$~\cite{uptime2024}---meaning that they use $1.56$ W of electricity for every $1$ W of compute.
This overhead is primarily spent on cooling, which can account for up to 40\% of the total power consumption of a data centre~\cite{capozzoli2015cooling}.
When estimating carbon emissions, this convolution of factors is compounded by the carbon intensity of the energy sources.
It can be particularly difficult to measure energy consumption in shared computing environments, such as a high-performance computing (HPC) cluster with multiple nodes, or virtualised hardware.
We use both of these classes of hardware in our experiments.
Our hardware is listed in Table~\ref{tab:machines}.

In this work, we focus on the energy usage of compute.
This energy can be measured using wall-mounted power meters~\cite{tschand2025mlperfpower} or estimated using software tools~\cite{Budennyy2022}. Although not as accurate as physical meters, these software tools can isolate the energy consumption of processes or devices, are more accessible and can be integrated with other software tools~\cite{Bouza2023}. For these reasons, we chose to build ECOpt around software-based energy meters. Specifically, we use CodeCarbon~\cite{courty2024codecarbon} for our experiments, as its readings have been found to be the most accurate and the most similar to those of wall-mounted meters~\cite{Bouza2023}. 

Software meters rely on the energy usage reported by the device drivers. For NVIDIA graphical processing units (GPUs), this is the NVIDIA System Management Interface (nvidia-smi), which has an error of $\pm 5\%$ in the reported power draw~\cite{Yang_2024}.
For x86-64 CPUs, this is the Running Average Power Limit (RAPL) interface. However, unprivileged access to the RAPL files under Linux was restricted in 2020, due to a security vulnerability\footnote{\url{https://github.com/mlco2/codecarbon/issues/244}}. Therefore, root access is typically required to measure the energy consumption of the CPU under Linux.

Measuring only the power consumption of the hardware accelerator, such as a GPU, is insufficient because different components are used at each stage of the ML workload.
In training, the GPU may be active during forward and backward propagation while the CPU is under load during data loading.
We list the power draw of each of our hardware components in Table~\ref{tab:machines-energy}.

In the context of ML inference, energy efficiency is often quantified by samples per Joule (J)~\cite{tschand2025mlperfpower}.
In keeping with convention and to reduce scale imbalance between our objective functions, we choose to adopt this energy efficiency metric, instead of samples per watt-hour (Wh).

To further support our choice to optimise for energy consumption, we conduct experiments in Section~\ref{sec:relationship} and Section~\ref{sec:exp-nas} to show that other cost metrics -- such as parameter and FLOP count -- can be uncorrelated with energy efficiency. However, ECOpt also supports other optimisation objectives, including time- and emissions-based metrics.

\subsection{Multi-objective optimisation}
\label{sec:impl-mobo}

As mentioned, we frame the task of quantifying the trade-off between performance and energy efficiency as an optimisation problem, employing multi-objective BO (MOBO) to solve it. This choice was inspired by future work posed by \citet{Husom2024}, and continues the line of practical applications of MOBO proposed by \citet{avent2020automatic} and \citet{ficiu2023automated}.

BO is a surrogate-based, sample-efficient global optimisation technique, which uses knowledge of prior samples to intelligently explore the search space~\cite{Mockus1975}. Given a function $g: \mathcal{X} \to \mathbb{R}$ of $n$ variables to optimise over a bounded domain with $\boldsymbol{x} = [x_1, \ldots, x_n] \in \mathcal{X} \subseteq \mathbb{R}^n$, BO searches for its minimum\footnote{The problem can be re-framed to search for the maximum.} as
\begin{displaymath}
    \boldsymbol{x}^* = \underset{\boldsymbol{x} \in \mathcal{X}}{\arg\min} \, g(\boldsymbol{x}).
\end{displaymath}

BO models the black-box function by fitting a surrogate model with sequentially acquired data. The surrogate model needs to be inexpensive to evaluate and able to quantify uncertainty over the various regions of the search space. In practice, the most common choice of surrogate model is a Gaussian process~\cite{williams2006gaussian}. BO uses an acquisition function $a: \mathcal{X} \to \mathbb{R}$ to balance the exploration (where the variance of the surrogate is high) and exploitation (where the mean of the surrogate is low) of the search space. A sample is taken where this heuristic function is maximised, at
\begin{displaymath}
    \boldsymbol{x}_t = \underset{\boldsymbol{x} \in \mathcal{X}}{\arg\max} \, a(\boldsymbol{x} \,|\, \mathcal{D}_{1:t-1}).
\end{displaymath}

\noindent The sample $g(\boldsymbol{x}_t)$ is added to $\mathcal{D}$ and used to update the surrogate function. This process is repeated until a stopping criterion is met.

MOBO is an extension of BO designed to optimise for multiple -- often conflicting -- objectives. Instead of searching for a single minimum, MOBO seeks to discover a Pareto frontier. A Pareto frontier is a surface in objective space that consists of Pareto-dominant points: a set of points in which no objective can be improved without sacrificing another. These points are generated by input-space solutions known as the Pareto set.

The solutions in the Pareto set dominate the other possible solutions.
We say that a solution $\boldsymbol{x} \in \mathcal{X}$ dominates $\boldsymbol{y} \in \mathcal{X}$ if $\boldsymbol{x}$ is at least as good as $\boldsymbol{y}$ in all objectives and strictly better in at least one objective. Formally, given $m$ objective functions $\boldsymbol{g} = [g_1, \ldots, g_m]: \mathcal{X} \to \mathbb{R}^m$, $\boldsymbol{x}$ dominates $\boldsymbol{y}$ if and only if $g_i(\boldsymbol{x}) \le g_i(\boldsymbol{y})$ for all $g_i \in \boldsymbol{g}$ and there exists $g_j \in \boldsymbol{g}$ such that $g_j(\boldsymbol{x}) < g_j(\boldsymbol{y})$.

We fit a separate Gaussian process model to each objective function, and assume homoscedastic additive zero-mean Gaussian noise levels. These models require a few initial observations before the iterative process of optimisation, which we collect at Sobol'~\cite{SOBOL196786} quasi-random points in the search space $S$.

The goal of MOBO is to find the Pareto set $\mathcal{P}$. This requires a specialised acquisition function that accounts for multiple objectives. A popular practical choice is expected hypervolume improvement (EHVI) \cite{emmerich2011ehvi}. EHVI estimates by how much a new solution would expand the dominated hypervolume of the Pareto frontier, given an anti-ideal reference point in the objective space. The anti-ideal point -- representing the worst anticipated value for each of the objective functions -- is used to calculate the hypervolume of the Pareto frontier and can be inferred using a quantile-based heuristic~\cite{mobo_tut}. In this work, we employ the parallel noisy EHVI ($q$NEHVI) \cite{Daulton2021} acquisition function, as it can account for our noisy measurements. To ensure that each objective contributes equally to the optimisation when calculating the hypervolume, we apply improvement normalisation and perform optimisation in the transformed objective space.

In ECOpt the MOBO procedure is implemented using Adaptive Experimentation Platform (Ax)~\cite{bakshy2018ae} and BoTorch~\cite{balandat2020}. For greater flexibility and reuse, we support multiple extensions to the procedure. For example, users can specify custom objective functions, provide fixed anti-ideal points and use pre-trained models.

\section{Experiments}
\label{sec:experiments}

\subsection{Overview}

This section covers results obtained using ECOpt. The experiments we report can be broadly divided into two groups: those that validate our design choices (Section \ref{sec:experiments-validation}) and those that demonstrate the effectiveness of our methodology by offering novel insights (Section \ref{sec:perf-evaluation}). All experiment tracking is performed using MLflow~\cite{zaharia2018accelerating}.

\subsection{Experimental setup}
\label{sec:exp-setup}
We begin by describing our experimental setup. The focus of our experiments is on deep learning models, which are driving the increase in the environmental impact of ML~\cite{2025Aquinoindex, Desislavov2023, Garcia-Martin2019}, and their inference energy consumption, because it is largely ignored by prior work~\cite{Desislavov2023}.
We primarily use image classification, since there are many public benchmarks for this task, but also consider text generation, due to the recent popularisation of large language models (LLMs).
The experiments are reproducible using the scripts in the provided source code.

\begin{table}
    \caption{The hardware specifications of the machines used in the experiments. The machines, including the virtual machine (VM), all run Linux. The server has two CPUs and eight GPUs, but we only use and measure one of each. Similarly, the HPC node has two CPUs, four GPUs and 1 TB of random-access memory (RAM) but we only request and measure the resources listed here. The VM only uses two of the CPU cores. Where applicable, we also list the available video RAM (VRAM) of the GPU.}\label{tab:machines}
    \begin{center}
        \resizebox{\columnwidth}{!}{
            \begin{tabular}{llrlr}
                \toprule
                Machine & CPU                             & RAM    & GPU                           & VRAM  \\
                \midrule
                VM      & Intel Xeon E5-2630 v4 & 8~GB   & ---                           & ---   \\
                Laptop  & Intel Core i7-1185G7            & 16~GB  & ---                           & ---   \\
                Desktop & Intel Core i7-10700             & 64~GB  & NVIDIA RTX 2080 & 8~GB  \\
                Server  & Intel Xeon Silver 4514Y         & 512~GB & NVIDIA L4                     & 24~GB \\
                HPC     & AMD EPYC 7763                   & 250~GB & NVIDIA A100                   & 80~GB \\
                \bottomrule
            \end{tabular}
        }
    \end{center}
\end{table}

\begin{table}
    \caption{The TDP of the components of the machines used in the experiments. The RAM TDP is based on the CodeCarbon estimate of 0.375~W/GB.}\label{tab:machines-energy}
    \begin{center}
        \begin{tabular}{lrlrrr}
            \toprule
            Machine & CPU                          & RAPL & RAM   & GPU   & Total \\
            \midrule
            VM      & $\frac{2}{10}(85)=17$ W & No   & 3~W   & ---   & 20~W  \\
            Laptop  & 28~W                         & Yes  & 6~W   & ---   & 34~W  \\
            Desktop & 65~W                         & No   & 24~W  & 250~W & 339~W \\
            Server  & 150~W                        & No   & 192~W & 72~W  & 414~W \\
            HPC     & 280~W                        & Yes  & 94~W  & 500~W & 874~W \\
            \bottomrule
        \end{tabular}
    \end{center}
\end{table}

Software optimisations and the choice of ML framework play an important role in the efficiency of model execution.
We use PyTorch~\cite{paszke2017pytorch}, both with our own model implementations and those downloaded from Hugging Face.
Where possible, we use small models to reduce the environmental impact of our research.
We use the default sampling rate and tracking mode of CodeCarbon in our experiments.
We observe that our measurements of energy consumption are fairly stable, with relative standard deviation not exceeding 1.5\%.
We perform our experiments using the machines listed in Table~\ref{tab:machines}, with power draws provided in Table~\ref{tab:machines-energy}.
The experiments -- including those failed or not presented -- have a combined running time of 39.80 hours, using 10.61~kWh of electricity and emitting 2.52~kgCO$_2$eq\footnote{Estimated by CodeCarbon based on the carbon intensity of the United Kingdom's energy mix in 2023.}.
This precise tracking of environmental impact, across distributed hardware, is enabled by ECOpt.

\subsection{Ablation experiments}
\label{sec:experiments-validation}

In this section, we report important ablation studies performed to validate our design choices. These experiments include investigating the relationship between model cost metrics and measuring the consistency of model energy consumption across hardware.

\subsubsection{Relationship between model cost metrics}
\label{sec:relationship}

There are various metrics for the computational cost of a model.
These can be arranged as a spectrum from abstract to hardware-specific: parameters, FLOPs or multiply-accumulate (MAC) operations, energy consumption, run-time and carbon emissions.
Previous work shows that there is a strong correlation between the number of parameters and the carbon emissions of a model~\cite{Luccioni2024powerhungry}.
\citet{Desislavov2023} find there to be a linear relation between the number of parameters and FLOPs of a model.
However, they observe that different architectures have different correlation coefficients; Transformers~\cite{vaswani2017attention} have a stronger coefficient than convolutional neural networks (CNNs).
In this experiment, we explore the relationship between the model cost metrics in neural networks and CNNs.
This experiment will inform whether the hardware-agnostic metrics of parameters or FLOPs can serve as proxies for energy consumption, or whether energy metrics instead need to be measured and published with models.

We observe the metrics of parameter count, energy consumption, FLOPs and run-time as we adjust model hyperparameters.
Since carbon emission is simply a function of energy consumption when the source of the energy is constant, we omit this metric from the plots in this experiment.
FLOPs and MACs are closely related; MACs count the number of floating-point multiplication and addition pairs (the dominant operation in neural networks) while FLOPs count all floating-point operations.
Therefore, the number of FLOPs is typically twice the number of MACs.
We choose to report the number of FLOPs, since this is a more general metric.
In particular, we use 32-bit FLOPs.
For this experiment, we use ECOpt's \verb|skip_train| argument to avoid training and logging of training metrics.
This saves time and energy, and places focus on the inference cost metrics.

\begin{figure}
    \centering
    \begin{subfigure}[t]{0.48\linewidth}
        \centering
        \includegraphics[width=\linewidth]{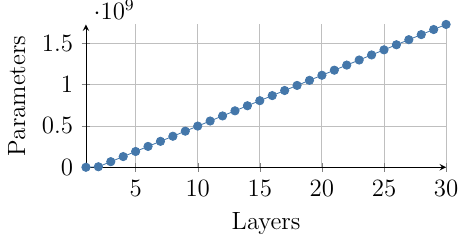}
        \caption{The parameter count per number of layers.}
    \end{subfigure}
    \hfill
    \begin{subfigure}[t]{0.48\linewidth}
        \centering
        \includegraphics[width=\linewidth]{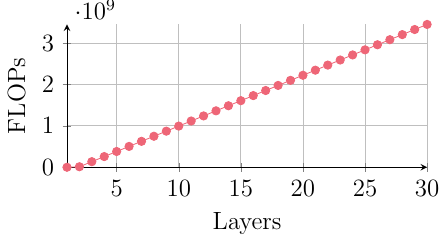}
        \caption{The FLOPs per number of layers.}
    \end{subfigure}\\
    \vspace{1em}
    \begin{subfigure}[t]{0.48\linewidth}
        \centering
        \includegraphics[width=\linewidth]{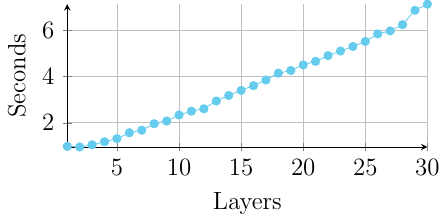}
        \caption{The run-time per number of layers.}
    \end{subfigure}
    \hfill
    \begin{subfigure}[t]{0.48\linewidth}
        \centering
        \includegraphics[width=\linewidth]{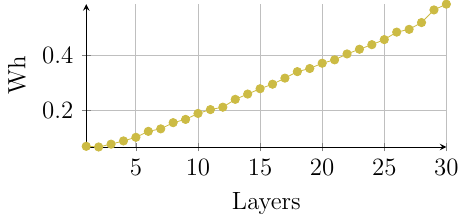}
        \caption{The energy usage per number of layers.}
    \end{subfigure}
    \caption{The cost metrics of the neural network model inference on 10,000 inputs in batches of 100, measured per number of layers on the server. We note that all the cost metrics grow linearly with the number of layers.}
    \Description{
    Panel (a): number of parameters is constant with regard to stride.
    Panel (b): number of FLOPs decreases with stride.
    Panel (c): run-time decreases with stride.
    Panel (d): energy usage decreases with stride.
    }
    \label{fig:exp-metrics-nn-server}
\end{figure}

We begin by evaluating these metrics for a simple neural network on 10,000 $28 \times 28$ inputs in batches of 100, as the number of layers is scaled from 1 to 30.
The network accepts inputs of size 784, has hidden layers of size 7,840 with rectified linear unit (ReLU)~\cite{nair2010rectified} activation functions, and an output layer of size 10.
In scaling this neural network from 7,850 to more than a billion parameters in Figure~\ref{fig:exp-metrics-nn-server}, we observe that all the cost metrics grow linearly with the number of layers. This is as expected.

We then experiment with a different model architecture; we implement a CNN parameterised by the hyperparameters listed in Table~\ref{tab:nas-hyperparams}.
It features an additional linear classification layer---scaled automatically to the output dimension of the CNN.
Each convolutional layer is followed by a RELU activation function and can optionally be followed by a two-dimensional max pooling layer.
This parameterised model later enables us to perform NAS in Section~\ref{sec:exp-nas}.

\begin{figure}
    \centering
    \begin{subfigure}[t]{0.48\linewidth}
        \centering
        \includegraphics[width=\linewidth]{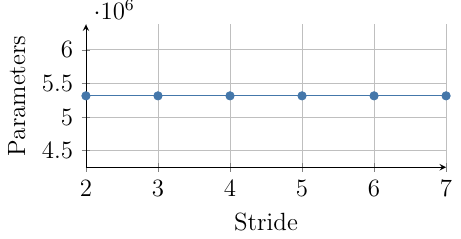}
        \caption{The parameter count per stride length.}
        \label{fig:exp-metrics-cnn-server-param}
    \end{subfigure}
    \hfill
    \begin{subfigure}[t]{0.48\linewidth}
        \centering
        \includegraphics[width=\linewidth]{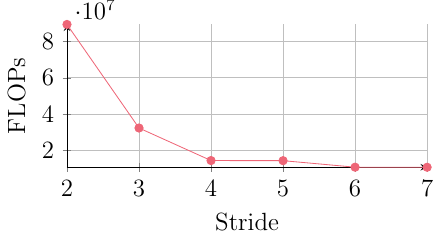}
        \caption{The FLOPs per stride length.}
        \label{fig:exp-metrics-cnn-server-flops}
    \end{subfigure}\\
    \vspace{1em}
    \begin{subfigure}[t]{0.48\linewidth}
        \centering
        \includegraphics[width=\linewidth]{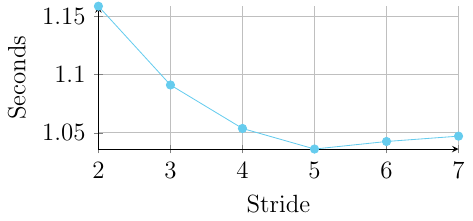}
        \caption{The run-time per stride length.}
    \end{subfigure}
    \hfill
    \begin{subfigure}[t]{0.48\linewidth}
        \centering
        \includegraphics[width=\linewidth]{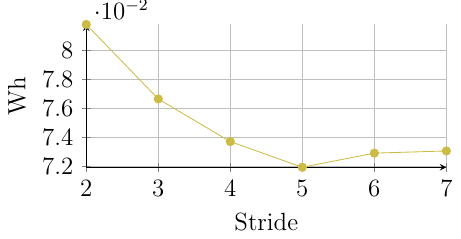}
        \caption{The energy usage per stride length.}
    \end{subfigure}
    \caption{The cost metrics of the CNN model inference on 10,000 inputs in batches of 1,000, measured per number of layers on the server. When scaling the stride, we observe that the number of parameters remains constant while the other cost metrics reduce.}
    \Description{
    Panel (a): number of parameters is constant with regard to stride.
    Panel (b): number of FLOPs decreases with stride.
    Panel (c): run-time decreases with stride.
    Panel (d): energy usage decreases with stride.
    }
    \label{fig:exp-metrics-cnn-server}
\end{figure}

We construct the model using 10 layers (without pooling) of 256 filters. The filters are of size three with a padding size of one.
We apply the model to the same 10,000 inputs as before, this time with a batch size of 1,000.
Figure~\ref{fig:exp-metrics-cnn-server} shows the effect of adjusting the stride length from two to seven.
We find that -- contrary to the results of \citet{Desislavov2023} -- the parameter count is not proportional to the other cost metrics.
This disparity is because stride does not affect the filters or their parameters but rather how they move across the input during convolution;
a larger stride skips over pixels, thus reducing the number of operations in Figure~\ref{fig:exp-metrics-cnn-server-flops}, but does not change the number of parameters in Figure~\ref{fig:exp-metrics-cnn-server-param}.
This result may explain the weaker correlation coefficient between parameter count and FLOPs of CNNs, observed by \citet{Desislavov2023}.

Mixture-of-Experts (MoE) is another example of a model architecture in which the number of parameters is not proportional to computational complexity~\cite{abnar2025parameters}.
We conclude that parameter count is an unsuitable cost metric for our objective function and should not be used as proxy for energy consumption.

The hardware is kept constant in this experiment, but \citet{tschand2025mlperfpower} observe a non-linear relationship between some cost metrics when changing hardware.
We next investigate the energy efficiency of models across hardware.

\subsubsection{Energy efficiency across hardware}
\label{sec:exp-consistency}

The power draw of ML systems can range from as low as 5.6~mW to at least 498.0~kW~\cite{tschand2025mlperfpower}.
Despite this, we hypothesise that the difference in energy efficiency across hardware configurations can be less than that between models.
\citet{2025Aquinoindex} measure the energy consumption of various image classification models on two different GPUs, finding a mere 5\% difference.
\citet{husom2024price} observe similar results in measuring the energy consumption of LLMs on two laptops and a desktop.
We conduct an investigation across a larger and more diverse range of hardware, listed in Table~\ref{tab:machines}, to test our hypothesis.
If our hypothesis is correct, then model energy efficiency is not wholly hardware-dependent and may be worth publishing.

We evaluate LLMs on the task of text generation, given its central role in the growing prominence of ML applications.
We choose recent Transformer~\cite{vaswani2017attention} models in a range of sizes, listed in Table~\ref{tab:consistency-models}.
These models are tasked with the same text generation task posed by \citet{Luccioni2024powerhungry}: generate 10 tokens in response to each of the first 1,000 entries of the BookCorpus dataset~\cite{zhu2015aligning}.
The entries are truncated to 20 words.
For determinism and fair comparison, inference is done without sampling and -- following the example of \citet{Luccioni2024powerhungry} -- we do not use batching.
For the VM, we limit the workload to the first 10 entries when running the two larger models. For the laptop, we do the same for the largest model.
This smaller sample size should not affect the chosen efficiency metrics and allow the experiment to be completed within feasible time.
Gemma 3 and Llama 3.1 cannot fit in the VRAM of the desktop and result in an out-of-memory (OOM) error.
For this reason, we do not provide results for these models on the desktop.

\begin{table}
    \caption{The text generation models used to experiment with the consistency of energy consumption across hardware. These models are chosen as recent Transformer models in a range of sizes.}\label{tab:consistency-models}
    \begin{center}
        \begin{tabular}{lrr}
            \toprule
            Model                                 & Parameters          & Published \\
            \midrule
            OpenAI GPT-2      & $132.00 \cdot 10^6$ & 2019      \\
            Alibaba Qwen3            & $752.00 \cdot 10^6$ & 2025      \\
            Google Gemma 3          & $4.00 \cdot 10^9$   & 2025      \\
            Meta Llama 3.1 & $8.03 \cdot 10^9$   & 2024      \\
            \bottomrule
        \end{tabular}
    \end{center}
\end{table}

\begin{figure}
    \centering
    \includegraphics[width=\linewidth]{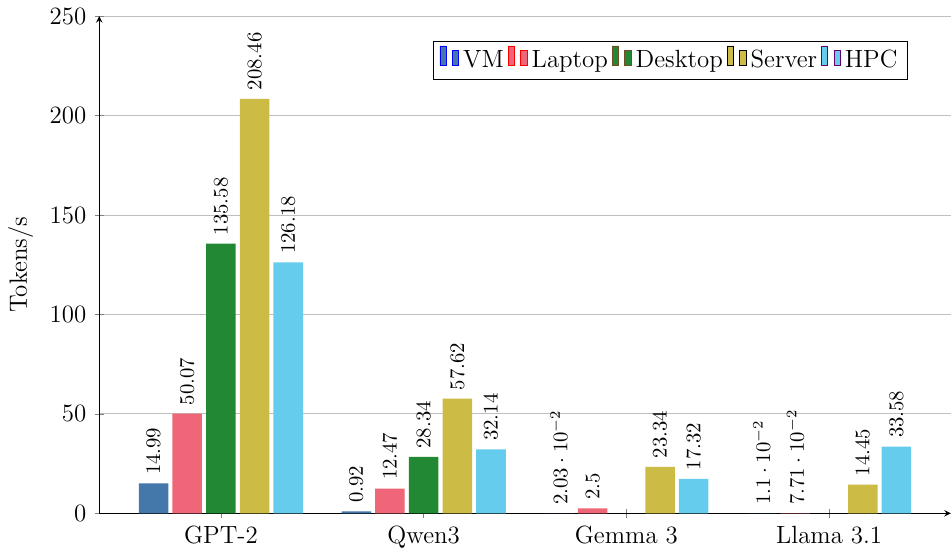}
    \caption{The throughput of each of the models in Table~\ref{tab:consistency-models} on the machines listed in Table~\ref{tab:machines}. Apart from the server outperforming the HPC for smaller models, the results are as expected: more powerful hardware produces more tokens per second. The desktop GPU cannot fit Gemma 3 or Llama 3.1 and is thus excluded from the results of these models.}
    \label{fig:consistency-per-second}
    \Description{Apart from the server outperforming the HPC for smaller models, the results are as expected: more powerful hardware produces more tokens per second. The desktop GPU cannot fit Gemma 3 or Llama 3.1 and is thus excluded from the results of these models.}
\end{figure}

As reference, we plot the tokens per second of each model on each machine in Figure~\ref{fig:consistency-per-second}.
One might expect that the smaller models and the more powerful machines exhibit greater performance.
However, this trend is broken by the HPC, which performs worse than the server when running the smaller models.
We ascribe this observation to the workload being input-output-bound for the smaller models. Therefore, the higher clock speed of the less powerful L4 GPU gives it an edge over the A100 GPU.

\begin{figure}
    \centering
    \includegraphics[width=\linewidth]{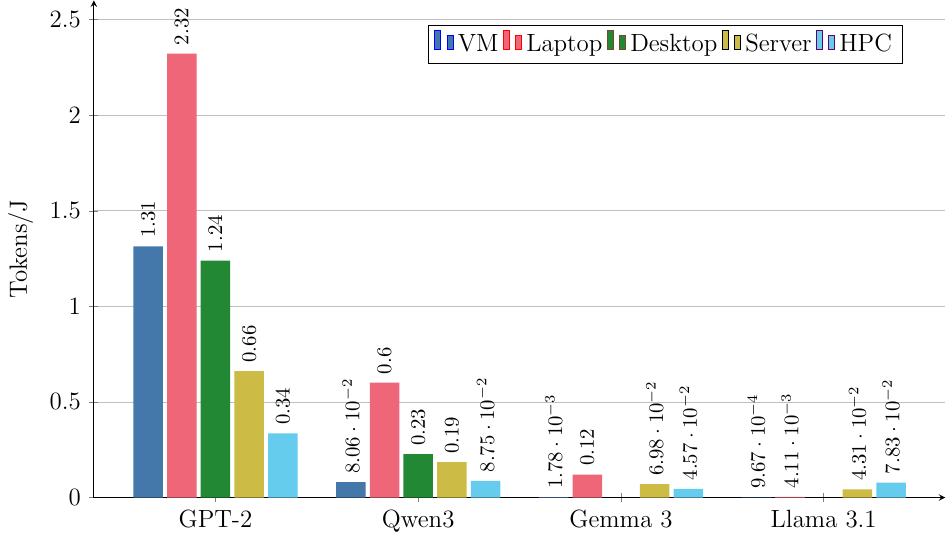}
    \caption{The energy efficiency of each of the models in Table~\ref{tab:consistency-models} on the machines listed in Table~\ref{tab:machines}. While there is a marked difference in energy efficiency between the models, they each show relatively consistent efficiency across machines. The CPU-based machines show a clear advantage for the sub-billion-parameter models, but perform worse with the larger models. Again, the desktop is excluded from the Gemma 3 and Llama 3.1 results due to limited VRAM.}
    \label{fig:consistency-per-joule}
    \Description{While there is a marked difference in energy efficiency between the models (more than four orders of magnitude on the laptop), they each show relatively consistent efficiency across machines (less than two orders of magnitude difference). The CPU-based machines show a clear advantage for the sub-billion-parameter models, but perform worse with the larger models. Again, the desktop is excluded from the Gemma 3 and Llama 3.1 results due to limited VRAM.}
\end{figure}

We now consider the energy efficiency of the models across the machines in Figure~\ref{fig:consistency-per-joule}.
The energy efficiency delta of each model across the machines is less than two orders of magnitude.
This is relatively consistent compared to the four-orders-of-magnitude difference between the energy efficiency of GPT-2 and Llama 3.1 on the VM.
To quantify this -- excluding the incomplete results of the desktop -- the standard deviation of the average energy efficiency of the models is 0.4608, compared to just 0.2372 for the machines.
Therefore, we consider it worth publishing model energy metrics, since they are relatively consistent across hardware in our experiments.

\begin{figure}
    \centering
    \includegraphics[width=\linewidth]{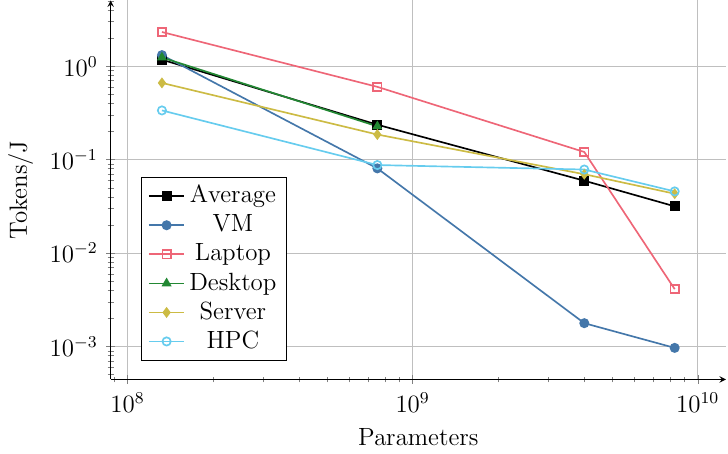}
    \caption{A log-log plot of the energy efficiency per number of parameters of the models and machines listed in Figure~\ref{fig:consistency-per-joule}. We observe a log-linear reduction in average energy efficiency as the models increase in size. Gemma 3 and Llama 3.1 are unable to fit into the desktop GPU memory.}
    \label{fig:efficiency-law}
    \Description{There is a log-linear reduction in average energy efficiency as the models increase in size. Gemma 3 and Llama 3.1 are unable to fit into the desktop GPU memory.}
\end{figure}

When plotting the energy efficiency against the Transformer model size in Figure~\ref{fig:efficiency-law}, we see that the energy efficiency scales predictably as a power law of the parameter count: decreasing log-linearly.
The marked drop in efficiency for the VM and laptop on the larger models is due to the models being unable to fit in the limited memory of these machines, resulting in page faults on the forwards pass.

The energy efficiency results are not all as expected.
Conventional wisdom says that GPUs are more efficient than CPUs for ML workloads~\cite{buber2018cpuvsgpu, li2016cpuandgpu}.
However, when considering GPT-2, the VM and laptop compute more tokens per joule than the GPU-accelerated machines.
The laptop continues to outperform these machines on the Qwen3 and Gemma 3 models—only falling behind on the larger Llama 3.1.
This observation aligns with the findings of \citet{Husom2024}, showing laptops to exhibit lower carbon emissions than a cloud server.
In our case, this is because the GPUs are not fully utilised by the smaller models; the server GPU compute only saturates with Llama 3.1, while the HPC only reaches 65\% GPU utilisation.
Not only does this waste the overhead energy required to power the GPUs, but it also does not take advantage of the more demanding hardware required to support the GPUs.
In the following experiment, we use ECOpt to optimise the workload to fully utilise the GPU.

\subsection{Performance evaluation}
\label{sec:perf-evaluation}

In this section, we demonstrate the effectiveness of our method. Specifically, we use ECOpt to optimise GPU utilisation for ML inference and perform an efficient neural architecture search on CIFAR-10.

\subsubsection{Optimising GPU utilisation}
\label{sec:exp-batch}

In the previous experiment, we observe that the GPU-accelerated hardware is not fully utilised when running the smaller models, such as Gemma 3 on the server.
Here, we show how ECOpt can fix this by optimising the batch size.
This is a simple use case, since the search space is one-dimensional and we only optimise for the efficiency objective, but it can demonstrate the efficacy of ECOpt in performing hardware-aware optimisation.

To focus optimisation on energy efficiency, we return a constant value of 100\% for the performance objective---making the hypervolume the identity function of the efficiency objective.
We further add a try-catch clause to handle OOM errors by returning an energy efficiency of zero tokens/J---penalising the model for exceeding the available VRAM.
We choose to run this experiment on the server instead of the HPC as the workload is too small to saturate the HPC GPU.
To facilitate fully utilising the GPU, we double the workload to the first 2,000 entries of BookCorpus and specify the batch size as a range hyperparameter of between 1 and 2,000.

\begin{figure}
    \centering
    \includegraphics[width=\linewidth]{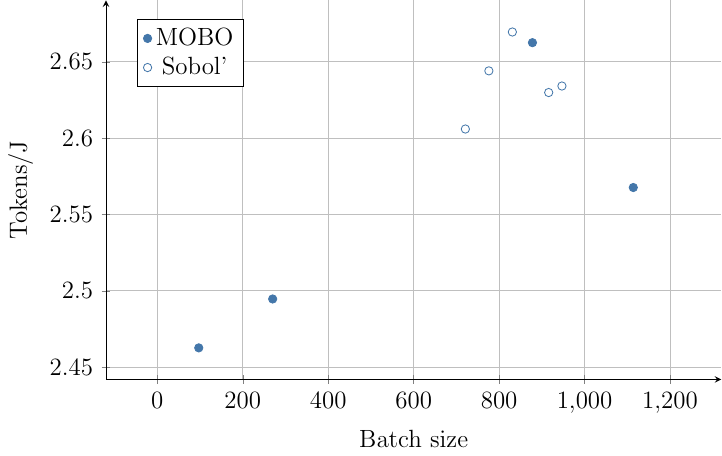}
    \caption{A scatter plot of the energy efficiency of Gemma 3 on the server per batch size, obtained using ECOpt with five Sobol' points and five MOBO steps. To preserve y-axis scale, one data point with batch size 1,374 that resulted in an OOM error has been hidden. The energy efficiency peaks at 2.67~tokens/J with a batch size of 831.}
    \label{fig:exp-batch}
    \Description{The energy efficiency peaks at 2.67~tokens/J with a batch size of 831.}
\end{figure}

Figure~\ref{fig:exp-batch} shows the energy efficiency measurements taken in applying the ECOpt optimiser to Gemma 3 on the server.
Interestingly, the optimal batch size is not the maximum batch size that fits into VRAM. ECOpt finds the optimal to be 831, with an efficiency of 2.67~tokens/J, a maximum GPU memory utilisation of 81\% and up to 100\% GPU compute utilisation.
This configuration represents a 38-fold increase in energy efficiency, compared to the default batch size of one, and it surpasses that of the laptop.
The experiment took a total of 264.30 seconds, using 18.61~Wh of energy.
We calculate that this 13.95 J reduction in energy consumption per token would break even with the energy used during optimisation after generating just 4,802 tokens.
This swift return on energy investment highlights the potential of ECOpt for positive environmental impact.

In the following experiment, we optimise for both energy efficiency and model performance, observing a trade-off between these metrics.

\subsubsection{Neural architecture search}
\label{sec:exp-nas}

Achieving higher accuracy with an ML model generally requires more computation and thus lower energy efficiency~\cite{tschand2025mlperfpower}.
In this experiment, we use ECOpt to perform NAS for a CNN model of CIFAR-10~\cite{krizhevsky2009learning}.
This showcases the utility of ECOpt and allows us to explore the compromise between energy efficiency and performance.
We conduct the experiment on the HPC to accelerate training while having access to the RAPL interface for accurate energy measurements.

The CIFAR-10 dataset contains 50,000 training images and 10,000 test images of $32 \times 32$ colour pixels.
The dataset is uniformly distributed over 10 image categories: `airplane', `automobile', `bird', `cat', `deer', `dog', `frog', `horse', `ship' and `truck'.
Due to this balanced distribution, we are able to use accuracy as the performance metric.
We normalise the data by subtracting the mean and dividing by the standard deviation of each colour channel.

We use the parameterised CNN model from Section~\ref{sec:relationship}.
While we choose to use a CNN for this experiment, it could be conducted using any model with a parameterised architecture.
For simplicity, we require that all the convolutional layers have the same dimensions and, to avoid dimensionality collapse with deeper networks, use a stride of one.
We keep the batch size and learning rate constant, since these would only serve to add noise to our NAS.
However, we show in Section~\ref{sec:exp-batch} how such hyperparameters can be optimised using ECOpt.

\begin{figure}
    \centering
    \includegraphics[width=\linewidth]{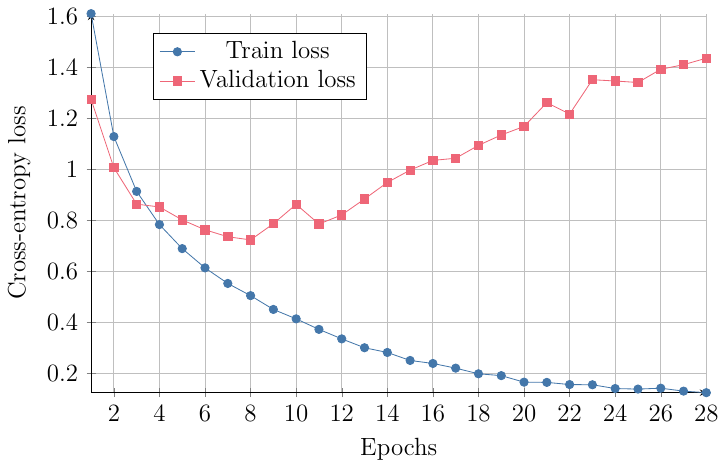}
    \caption{The training and loss curves of the CNN (with six layers, pooling enabled, 51 filters and a kernel size of three) on CIFAR-10 for 28 epochs. The divergence in train and validation loss suggests over-fitting to the training data after four epochs and motivates our use of early stopping.}
    \label{fig:nas-loss}
    \Description{The divergence in train and validation loss suggests over-fitting to the training data after four epochs and motivates our use of early stopping.}
\end{figure}

Since the candidate architectures vary in parameter count, they reach training saturation at different points.
Therefore, some of the models would over-fit while others would under-fit to the training data, if we used a fixed number of epochs for all the models.
An example of this over-fitting is shown in Figure~\ref{fig:nas-loss}.
To prevent this, we implement early stopping of training, with an upper bound of 500 epochs.
Early stopping is triggered once the validation loss delta drops below 0.001 for three consecutive epochs.
To calculate validation loss, we pseudo-randomly reserve 10,000 of the training images as validation set.
This splitting of the training set further allows for fair comparison with the results of \citet{aszemi2019hyperparameter}, who also reserve 10,000 training images for validation and train for 500 epochs with early stopping.

\begin{figure}
    \centering
    \includegraphics[width=\linewidth]{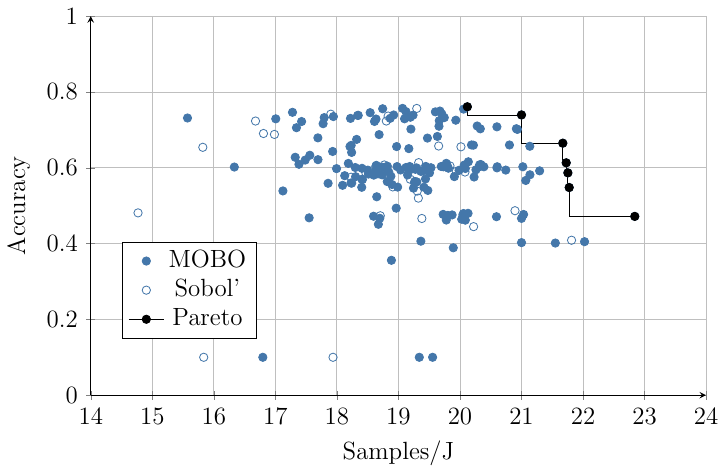}
    \caption{The Pareto frontier of the CNN NAS, identifying seven Pareto-optimal points from a pool of 40 Sobol' and 160 MOBO points. The search is conducted on the HPC with the CIFAR-10 dataset. The frontier shows a clear trade-off between energy efficiency and performance. The six models that score exactly 10\% accuracy always predict the same class. This over-fitting is due to the prevalence of this class in the training dataset after the validation images are pseudo-randomly removed.}
    \label{fig:nas-pareto}
    \Description{The frontier shows a clear trade-off between energy efficiency and performance.}
\end{figure}

For training, we use cross-entropy loss with the Adam optimiser, a learning rate of 0.001 and a batch size of 64, as this configuration is found to perform well on CIFAR-10 with small CNNs like ours~\cite{aszemi2019hyperparameter, lee2025investigating}.
Table~\ref{tab:nas-hyperparams} summarises the hyperparameters and their domains.
These generate a four-dimensional search space of $6 \times 2 \times 128 \times 5 = 7{,}680$ possible architectures.
We apply the ECOpt optimiser to this model, sampling 40 Sobol' points before generating 160 candidate architectures using MOBO.
This process enables us to visualise the compromise between model performance and energy efficiency in Figure~\ref{fig:nas-pareto}, by identifying a Pareto set of seven architectures.
We find no single architecture that scores highest in both metrics.

The plot shows ECOpt to focus candidate generation in the elbow of the Pareto frontier, where there are gains to be made in both metrics.
Since we do not specify threshold values, these are inferred as explained in Section~\ref{sec:impl-mobo}.
For this reason, we see few candidates with less than 40\% accuracy.
If we wished to further explore the Pareto frontier, we could specify an accuracy threshold of 0\%.
Similarly, there are no models that achieve above 80\% accuracy.
This is due to a limitation of our architecture; the most computationally expensive model in the search space is only able to achieve 48.63\% accuracy.
For the performance objective, the six outlying models all score exactly 10\% accuracy: predicting the same class irrespective of the input.
This over-fitting is due to an imbalance in the training data caused by the pseudo-random removal of the validation set.

\begin{table}
    \caption{The CNN hyperparameters optimised by ECOpt in the NAS experiment. The max pool hyperparameter specifies whether or not to add a two-dimensional max pooling layer with a stride and kernel size of two after every convolutional layer.}\label{tab:nas-hyperparams}
    \begin{center}
        \resizebox{\columnwidth}{!}{
            \begin{tabular}{lllr}
                \toprule
                Hyperparameter       & Type   & Data type      & Domain                \\
                \midrule
                Layers               & Range  & Integer        & [1, 6]                \\
                Max pool             & Choice & Boolean        & \{True, False\}       \\
                Filters              & Range  & Integer        & [1, 128]              \\
                Kernel size          & Choice & Integer        & \{1, 3, 5, 7, 9\}     \\
                Padding              & Fixed  & Integer        & (Kernel size - 1) / 2 \\
                Stride               & Fixed  & Integer        & 1                     \\
                Epochs               & Fixed  & Integer        & 500                   \\
                Batch size           & Fixed  & Integer        & 64                    \\
                Learning rate        & Fixed  & Floating-point & 0.001                 \\
                Stop early           & Fixed  & Boolean        & True                  \\
                Stop early patience  & Fixed  & Integer        & 3                     \\
                Stop early min delta & Fixed  & Floating-point & 0.001                 \\
                \bottomrule
            \end{tabular}
        }
    \end{center}
\end{table}

\citet{aszemi2019hyperparameter} perform hyperparameter tuning of CNNs similar to ours on CIFAR-10, using random search.
Their tuning takes three days using an NVIDIA Tesla K80 with 300~W TDP, achieving 71.17\% accuracy on the test set.
Using ECOpt, we achieve 76.09\% accuracy in this experiment, taking just 61 minutes.
This efficiency is due to the guided candidate generation of MOBO.
We also improve upon the 70.07\% accuracy achieved by \citet{lee2025investigating} in performing manual hyperparameter tuning of CNNs on CIFAR-10.
Furthermore, we find a range of suitable architectures for the user to select at their desired level of energy efficiency or accuracy.
These are listed in Table~\ref{tab:nas-optimal}.

\begin{table}
    \caption{The metrics and hyperparameters of Pareto-optimal architectures identified in the CNN NAS experiment.}\label{tab:nas-optimal}
    \begin{center}
        \resizebox{\columnwidth}{!}{
            \begin{tabular}{rrrlrr}
                \toprule
                Accuracy & Samples/J & Layers & Max pool & Filters & Kernel size \\
                \midrule
                76.09\%  & 20.12     & 4      & True     & 83      & 3           \\
                73.94\%  & 21.00     & 4      & True     & 50      & 3           \\
                66.49\%  & 21.67     & 3      & True     & 26      & 7           \\
                61.31\%  & 21.73     & 1      & False    & 40      & 3           \\
                58.66\%  & 21.75     & 1      & False    & 16      & 3           \\
                54.79\%  & 21.78     & 1      & True     & 98      & 9           \\
                47.17\%  & 22.84     & 4      & False    & 31      & 1           \\
                \bottomrule
            \end{tabular}
        }
    \end{center}
\end{table}

Although our simple models are far from SOTA accuracy on CIFAR-10, they could be SOTA in terms of accuracy/J.
However, in Section~\ref{sec:related}, we argue against collapsing energy efficiency and performance metrics to a single measure.
The current SOTA in both energy consumption and accuracy on CIFAR-10 is the Spike Aggregation Transformer (SAFormer)~\cite{zhang2024combining}.
It has an accuracy of 95.80\% on CIFAR-10 and theoretical energy consumption of 0.49~mJ per inference.
SAFormer is specifically designed for this theoretical energy consumption measure, which is based on the assumption that MAC operations require 4.6~pJ to compute.
By this measure, our model that achieves 76.09\% accuracy and requires 85,637,408 MACs has an energy consumption of just 0.39~mJ.
While our model is less accurate, it is more energy-efficient and, therefore, not dominated by SAFormer.
Since this is the least energy-efficient of our identified architectures, all seven fall on the Pareto frontier of accuracy and energy efficiency for CIFAR-10.

While evaluating the MOBO candidate architectures, we compare the energy consumption to parameter count, FLOPs and run-time.
As in Section~\ref{sec:relationship}, we observe that the parameter count is not correlated with energy consumption during inference.
We find that is has a Pearson's correlation coefficient~\cite{pearson1895vii} of -0.12.
This time, however, stride is kept constant.
We attribute this uncoupling of metrics to the use of pooling, which reduces the workload without affecting the parameter count.
Furthermore, we see that FLOPs is also an unreliable cost metric for CNNs, as -- with a coefficient of 0.26 -- there is no clear correlation between it and energy consumption.
We postulate that this is as a result of the smaller models not fully utilising the GPU.
Other than carbon emissions, time is the only metric to be strongly correlated with energy consumption in this experiment; run-time has a correlation coefficient of 0.79 with energy consumption. \citet{nguyen2025on-deviceorremote} also find a statistically significant positive correlation between run-time and energy consumption of on-device model inference.

\citet{eriksson2021latency} find that wider, shallower architectures are preferable when optimising for accuracy and latency on mobile devices.
In Table~\ref{tab:nas-optimal}, we list the hyperparameters and metrics of the optimal points found in the NAS.
We observe a mix of shallow architectures without pooling and deeper models with pooling.
This observation highlights the difference in optimising for latency and for energy efficiency.
Both objectives require minimising compute time.
However, optimising for latency calls for maximising the utilisation of hardware with wide networks, while optimising for energy efficiency requires minimising power draw by balancing width and depth.

\section{Threats to validity}

\subsection{Internal threats}

\paragraph{Energy measurement accuracy}
Due to access restrictions to the RAPL interface under Linux, our experiments on the VM, desktop, and server machines rely on estimated CPU energy consumption rather than direct hardware measurements. Similarly, recent work has shown CodeCarbon measurements to be off by up to 40\%~\cite{fischer2025groundtruthingaienergyconsumption}. Although this prevents us from reporting absolute energy consumption values with high precision, it does not compromise the validity of relative comparisons between successive measurements within the same experimental setup. This threat is mitigated by the consistency of our estimation approach in all comparative evaluations.

\paragraph{Incomplete energy accounting}
Following \citet{tschand2025mlperfpower}, we measure only the energy used for compute and exclude the energy used for cooling. Isolating the energy consumption attributable to cooling a specific node within a data centre presents significant technical challenges. However, ECOpt allows users to specify a PUE value for their hardware configuration. This enables them to approximate the total energy consumption for their specific context.

\paragraph{Energy measurement scope}

The hardware power monitoring interfaces that CodeCarbon uses aggregate package power (including compute units, memory controllers, on-chip caches and data transfers) rather than provide isolated computational energy. Although we cannot decompose measurements to isolate data storage and movement energy from compute energy, our experiments implicitly capture these effects through aggregate measurements.

\subsection{External threats}

\paragraph{Inference optimisation techniques}
Our evaluation does not incorporate common inference optimisations such as query caching~\cite{zhu2023optimal} and model routing~\cite{salmani2023}. These techniques can significantly impact energy consumption in production deployments. Although this limits the direct applicability of our absolute results to production systems employing these techniques, our approach remains valid and can be applied to models that incorporate such optimisations. Moreover, ECOpt provides a convenient method of implementing these techniques, via the \verb|evaluate| function of the model wrapper class.

\section{Conclusion}
\label{sec:conclusion}

In this work, we describe ECOpt: a tool to measure the energy efficiency of any ML model, optimise its hyperparameters, visualise the trade-off between energy efficiency and performance, and track experiments.
ECOpt can enable ML practitioners to maximise the performance of their models while reducing their energy cost and environmental impact.
It can also help them comply with new regulations, such as the EU AI Act.

In our experiments using ECOpt, we explore the relationship between metrics for the computational cost of a CNN model and find that parameter count is an unreliable proxy.
In measuring the energy efficiency of Transformer models for text generation, we observe that it is relatively consistent across hardware, and thus worth publishing.
We observe that CPUs can be more energy efficient for ML workloads when GPUs are not properly utilised and uncover an energy scaling law in Transformer models: energy efficiency decreases log-linearly with the number of parameters.

By applying ECOpt to Gemma 3, we find that the optimal batch size in terms of energy efficiency does not necessarily saturate the VRAM.
Furthermore, we show that ECOpt is able to recuperate the energy spent during optimisation in as few as 4,802 tokens.
Using ECOpt to perform NAS for CNN models of CIFAR-10, we observe a compromise between energy efficiency and accuracy.
We demonstrate ECOpt to improve upon existing hyperparameter tuning methods in terms of both efficacy and efficiency.
In this experiment, we show that FLOPs can also be uncorrelated with the energy cost of CNNs, and discuss how the goals of optimising for latency differ from those of optimising for energy efficiency.

Using ECOpt, we find seven models for CIFAR-10 that improve SOTA when considering accuracy and energy efficiency together.
However, this was not our intention and is largely due to the scarcity of published energy metrics.
We hope that others will use ECOpt to develop models that dominate ours.

There are several exciting directions to extend this work. For example, MOBO is known to scale poorly with the number of input parameters. Therefore, it might be necessary to explore techniques that extend MOBO to higher dimensions -- such as sparse axis-aligned subspace~\cite{eriksson2021high} -- to optimise models with many hyperparameters. Another promising direction is to apply ECOpt to distributed ML techniques, since techniques such as federated learning have been shown to produce up to two orders of magnitude more carbon emissions than centralised learning~\cite{qiu2023first}. Finally, we would like to validate our methodology by using ECOpt to optimise large-scale models in close-to-production environments.

We encourage ML practitioners and researchers to publish the energy efficiency of their models to foster competition and promote awareness of the environmental impact of ML. We discuss relevant metrics in Section~\ref{section:energy-measure} and invite readers to take inspiration from existing initiatives, such as Hugging Face's AI Energy Score\footnote{\url{https://huggingface.github.io/AIEnergyScore}}. Moreover, reducing the cost of training and deployment through more efficient models could make the industry more accessible to a wider range of participants by lowering financial barriers to entry.


\bibliographystyle{ACM-Reference-Format}
\bibliography{ecopt}










\end{document}